\documentclass[conference]{IEEEtran}
\usepackage{cite}
\usepackage{parskip}
\usepackage{graphicx}

\begin{document}
\title{Is AmI (Attacks Meet Interpretability) \\
  Robust to Adversarial Examples?\vspace{-1em}}
\author{\IEEEauthorblockN{Nicholas Carlini (\emph{Google Brain})}}
\maketitle

\begin{abstract}
  No.% AmI does not offer meaningful robustness.
  \end{abstract}

\section{Attacking ``Attacks meet Interpretability''}

AmI (Attacks meet Interpretability) is an
 ``attribute-steered'' defense \cite{tao2018attacks} to
detect \cite{carlini2017adversarial} adversarial examples \cite{szegedy2013intriguing}
on face-recognition models.
By applying interpretability techniques to a
pre-trained neural network, AmI identifies
``important'' neurons.
It then creates a second augmented neural network with the same
parameters but increases the weight activations of important 
neurons. AmI rejects inputs where the original and augmented
neural network disagree.% An adversarial example is therefore one
%where \emph{both} the original and augmented network agree on the
%\emph{same} wrong output.

We find that this defense (presented at at NeurIPS 2018 as
a spotlight paper---the top 3\% of submissions)
is completely ineffective, and even \emph{defense-oblivious}%
\footnote{We mount a defense-oblivious attack because it shows that
  even under this incredibly weak threat model the defense is ineffective. (The defense is also written in Caffe which the author
  did not want to have to use.)
Future defenses should \emph{not} argue security only under this threat model.}
attacks reduce the detection rate to $0\%$ on untargeted attacks.
That is, AmI is no more robust to untargeted attacks than the undefended original network.
Figure~1 contains examples of
adversarial examples that fool
the AmI defense.
We are incredibly grateful to the authors for releasing their
source code%
\footnote{https://github.com/AmIAttribute/AmI}
which we build on%
\footnote{https://github.com/carlini/AmI}. We hope that
future work will continue to release source code by
publication time to accelerate progress in this field.

\subsection{Evaluation}

We assume familiarity with prior work (specifically
\cite{carlini2017adversarial,szegedy2013intriguing,tao2018attacks}).

Unfortunately, the defense paper \cite{tao2018attacks} does
not contain a threat model or
make any \emph{specific} claims about robustness,
making it difficult to
perform a proper security evaluation.
The authors, in personal communication, 
stated the bound was meant to be $0.01$ ($3\times$ \emph{lower}
than is in almost all other prior work);
this extremely low distortion bound is never given in the paper.

We generate adversarial examples by completely ignoring the defense and
generating high-confidence adversarial examples on the original neural
network. This approach, while simple, has proven
surprisingly successful in the past when
attacking detection-based defenses \cite{carlini2017adversarial}.
We choose an (incorrect) target label at
random and generate a high-confidence targeted adversarial example
for that target using \emph{only} the original network. We then
test to see if the resulting image happens by chance to be
adversarial on the combined
defended model (i.e., is misclassified the same way by both
networks). If it is not (and would therefore be rejected),
we repeat the process and try again
until we succeed. The median number of attempts is 25.

This na\"ive attack is successful $100\%$ of the time:
the detector has a $0\%$ true-positive rate
(\emph{lower} than the $9.9\%$ false positive rate);
Figure~1 contains successful adversarial examples.

\section{Conclusion}
``Attacks Meet Interpretability'' \cite{tao2018attacks}
is not robust to untargeted adversarial examples 
with $\ell_\infty$ bound $0.01$, even when the attacker is
oblivious to existence of the defense.
While our attack is not efficient, we believe an adaptive
attack
that specifically targeted the defense would be
much more efficient while remaining $100\%$ successful at evading
detection.

We implore researchers who
propose defenses to investigate \emph{why}
attacks fail before declaring a proposed defense effective;
and similarly implore those reading or reviewing
papers to think critically about why the attacks could have
failed before believing the claimed defense results.
It is exceptionally easy to fool oneself when evaluating adversarial
example defenses, and every effort must be taken to ensure that
when attacks fail it is not
because attacks have been applied incorrectly.

\begin{figure}[t]
  \centering
  \vspace{1em}
  \includegraphics[scale=.4]{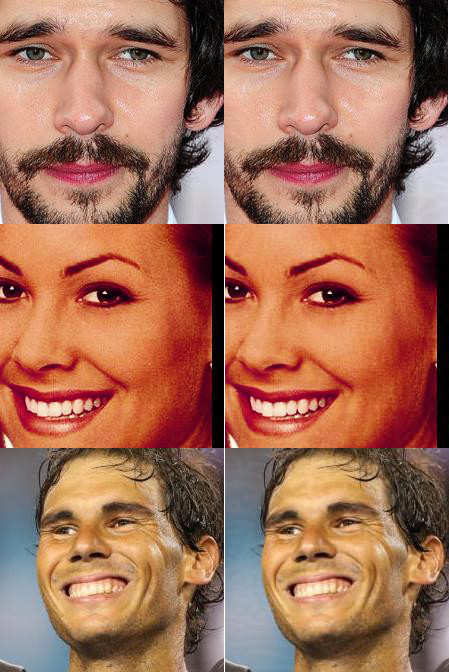}
  \vspace{-.5em}
  \caption{(left) Original images; (right) adversarial examples defeating AmI.}
\end{figure}
\vspace{-.5em}
{\footnotesize
\bibliographystyle{IEEEtranS}
\bibliography{paper}
}

\end{document}